\documentclass[10pt,twocolumn,letterpaper]{article}
\usepackage{cvpr} 

\usepackage[belowskip=-5pt,aboveskip=3pt]{caption} 
\usepackage[labelfont=bf]{caption}
\usepackage{listings}
\usepackage{array}
\usepackage{multirow}
\usepackage{multicol}
\usepackage{tabularx}
\usepackage{dsfont}
\usepackage{url}
\usepackage{color}
\usepackage{caption}
\usepackage{amsthm}
\usepackage[export]{adjustbox}
\usepackage{comment}

\usepackage[belowskip=1pt,aboveskip=0pt]{caption}

\usepackage[utf8]{inputenc} 
\usepackage[T1]{fontenc}    
\usepackage{url}            
\usepackage{booktabs}       
\usepackage{amsfonts}       
\usepackage{nicefrac}       
\usepackage{microtype}      
\usepackage{enumitem}
\usepackage{fancyhdr}
\usepackage{fancyhdr}

\def\argmax{\operatornamewithlimits{arg\,max}}

\usepackage{pifont}
\newcommand{\reqref}[1]{Eq.~\eqref{#1}}

\usepackage{colortbl}
\usepackage{times}
\usepackage{epsfig}
\usepackage{graphicx}
\usepackage{amsmath}
\usepackage{amssymb}
\usepackage{booktabs}

\usepackage{epsfig}
\usepackage{graphicx}
\usepackage{amsmath}
\usepackage{enumitem}
\usepackage{algorithm}
\usepackage{algorithmic} 
\usepackage{listings}
\usepackage{makecell}
\usepackage{xspace}

\makeatletter
\DeclareRobustCommand\onedot{\futurelet\@let@token\@onedot}
\def\@onedot{\ifx\@let@token.\else.\null\fi\xspace}
\def\eg{\emph{e.g}\onedot} 
\def\ie{\emph{i.e}\onedot} 
 
\def\etc{\emph{etc}\onedot} 
 
\def\etal{\emph{et al}\onedot}
\makeatother

\newcommand\Statex{\Statex\hspace{\algorithmicindent}}

\AtBeginDocument{%
  \providecommand\BibTeX{{%
    \normalfont B\kern-0.5em{\scshape i\kern-0.25em b}\kern-0.8em\TeX}}}
    
\author{Huiming Sun\textsuperscript{1}, Lan Fu\textsuperscript{2}, Jinlong Li\textsuperscript{1}, Qing Guo\textsuperscript{3},\\ Zibo Meng\textsuperscript{2}, Tianyun Zhang\textsuperscript{1}, Yuewei Lin\textsuperscript{4}, Hongkai Yu\textsuperscript{1}\\ \\
\textsuperscript{1}Cleveland State University. \textsuperscript{2}OPPO US Research Center. \\
\textsuperscript{3}A*STAR SINGA. 
\textsuperscript{4}Brookhaven National Laboratory. 
}

\begin{document}

\title{Defense against Adversarial Cloud Attack on Remote Sensing \\ Salient Object Detection}
\maketitle

\begin{figure*}[th]
    
\centering
\footnotesize
  \includegraphics[width=\textwidth]{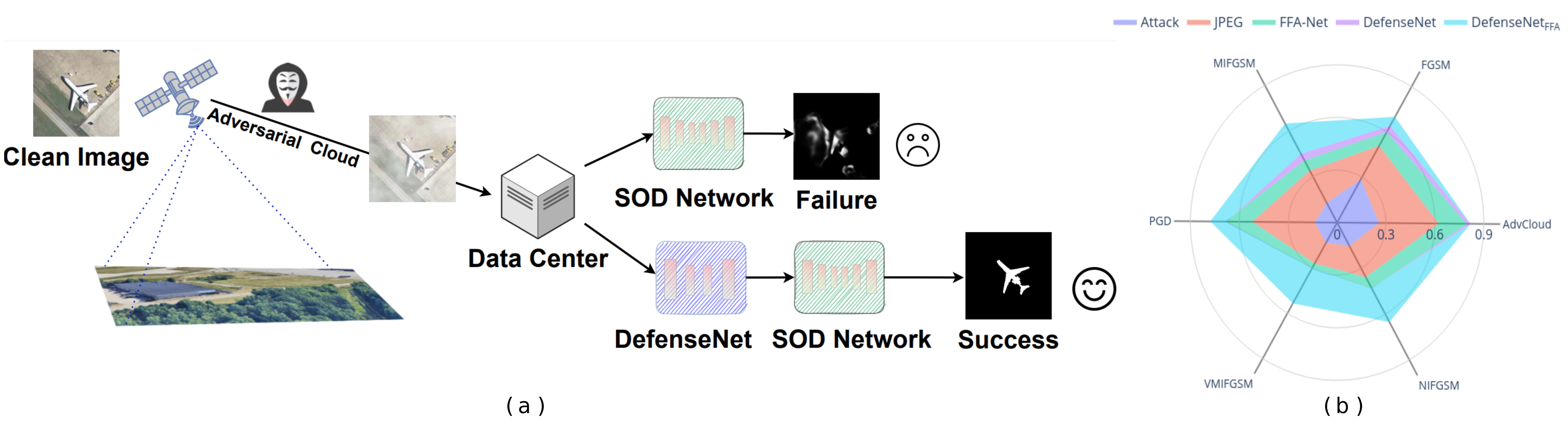}
  \caption{(a) Illustration of the proposed defense against the adversarial cloud attacks for remote sensing salient object detection. (b) Performance (F Measure) of the proposed DefenseNet against different adversarial cloud attacks. Bigger area means better defense.}
  \label{fig:motivation}
\end{figure*}

\begin{abstract}
 
Detecting the salient objects in a remote sensing image has wide applications for the interdisciplinary research.  
Many existing deep learning methods have been proposed for Salient Object Detection (SOD) in remote sensing images and get remarkable results. 
However, the recent adversarial attack examples, generated by changing a few pixel values on the original remote sensing image, could result in a collapse for the well-trained deep learning based SOD model.
Different with existing methods adding perturbation to original images, we propose to jointly tune adversarial exposure and additive perturbation for attack and constrain image close to cloudy image as Adversarial Cloud. 
Cloud is natural and common in remote sensing images, however, camouflaging cloud based adversarial attack and defense for remote sensing images are not well studied before.  
Furthermore, we design DefenseNet as a learn-able pre-processing to the adversarial cloudy images so as to preserve the performance of the deep learning based remote sensing SOD model, without tuning the already deployed deep SOD model.
By considering both regular and generalized adversarial examples, the proposed DefenseNet can defend the proposed Adversarial Cloud in white-box setting and other attack methods in black-box setting. 
Experimental results on a synthesized benchmark from the public remote sensing SOD dataset (EORSSD) show the promising defense against adversarial cloud attacks. 
\end{abstract}






\section{Introduction}
The cross-domain research of computer vision and remote sensing has wide applications in the real world,  such as hyperspectral image classification~\cite{wang2016salient, zhang2018diverse},  cross-view geolocation~\cite{tian2017cross, zhu2021revisiting},  scene classification~\cite{zou2015deep, song2019domain},  change detection~\cite{bruzzone2002adaptive, du2019unsupervised},  aerial-view object detection~\cite{Xia_2018_CVPR, ding2019learning},  and so on. Salient Object Detection (SOD) in remote sensing images is to extract the  salient objects in a satellite or drone image,  which might benefit many research works mentioned above. 



Some existing methods have been proposed for the SOD task in remote sensing images~\cite{li2019nested, zhang2020dense} using Convolutional Neural Network (CNN) based network architecture, whose efforts are mainly focused on multi-scale feature aggregation~\cite{li2019nested} and representative context feature learning~\cite{zhang2020dense}. However, in some scenarios, these deep learning based remote sensing SOD models might suffer from the attacks by the  adversarial examples on deep neural networks. Recent research~\cite{gao2022can} shows that the adversarial noises can be added to fool the deep learning based SOD models, leading to the low SOD performance. For example, by adding a small portion of adversarial noises on the original remote sensing image between the image acquisition and data processing, \eg, during the communication, the salient objects in the remote sensing image might  be hided or missed to some extents by the deep SOD model. This kind of malicious attack exposes a potential security threat to the remote sensing.

Many researches have been proposed for the adversarial examples based attack and defense in deep  learning~\cite{goodfellow2014explaining,gao2021advhaze,liao2018defense,zhang2019defense}. Meanwhile, some attack and defense researches have been proposed for remote sensing tasks, such as the remote sensing scene classification~\cite{xu2020assessing}. Different with existing methods adding the perturbation on the original image, we propose to generate Adversarial Cloud as attack to the deep learning based remote sensing SOD model. Cloud is widely common in remote sensing images~\cite{sun2021convolutional}. However, cloud based adversarial attack and defense for remote sensing images has not been well studied. The proposed Adversarial Cloud has realistic appearance close to a normal cloud, which might be difficult to be perceived but will be malicious in the remote sensing applications.

In this paper, we propose a novel DenfenseNet to defend the proposed Adversarial Cloud attack to preserve the advanced SOD performance. In general, the adversarial attack and defense networks will be trained with an adversarial deep learning by iteratively training the Adversarial Cloud and DenfenseNet. However, the already deployed deep remote sensing SOD model is kept unchanged to simplify the real-world setting. Thus, the proposed DefenseNet is designed as a learn-able pre-processing technique to preserve the SOD performance. In specific, the adversarial examples will go through the DefenseNet to become clean examples as the input to SOD models. Based on the   publicized remote sensing SOD dataset (EORSSD~\cite{zhang2020dense}), we build a benchmark by synthesizing the Adversarial Cloud to test the performance of attack and defense for the SOD problem in the remote sensing images. As shown in Fig.~\ref{fig:motivation}~(b), our proposed method could defend different adversarial attack methods. Experimental results on the built benchmark show the effectiveness and accuracy of the proposed method. The contributions of this paper are summarized as follows. 

\begin{itemize}[noitemsep, nolistsep,leftmargin=*] 
    \item This paper proposes a novel attack  method by jointly tuning adversarial exposure and additive perturbation and constraining image close to cloudy image as Adversarial Cloud for the SOD in remote sensing images. 
    
    \item This paper proposes a novel DefenseNet as learn-able pre-processing against the adversarial cloud attack for the safety-ensured SOD in remote sensing images, without tuning the already deployed deep learning based SOD model. 
    
    \item By considering both regular and generalized adversarial examples, the proposed DefenseNet can defend the proposed Adversarial Cloud in white-box setting and other attack methods in black-box setting.
\end{itemize}

\vspace{-5pt}
\section{Related Work}
\subsection{Salient Object Detection for Remote Sensing}
Salient object detection (SOD) is to automatically extract the salient objects in an image. Many existing methods have been proposed for SOD in natural images,  while the SOD in optical remote sensing images is more challenging due to the unique,  complex and diverse environments~\cite{li2019nested}. 
SOD in satellite or drone images has wide applications in remote sensing,  such as building extraction~\cite{li2016building},  Region-of-Interest extraction~\cite{ma2016region},  airport detection~\cite{zhang2018airport},  oil tank detection~\cite{liu2019unsupervised},  ship detection~\cite{dong2019ship},  \etc.

Some traditional methods have been proposed for SOD in remote sensing images by employing the bottom-up SOD models~\cite{zhao2015sparsity, zhang2019saliency, ma2016region, zhang2013region, zhang2015region}. Recently,  more deep learning based SOD methods are proposed for the optical remote sensing  images~\cite{li2019nested, li2020parallel, zhang2021salient, zhang2020dense, cong2021rrnet}. The efforts of these deep learning based methods are mainly focused on multi-scale feature  aggregation,  \eg,  ~\cite{li2019nested} and representative context feature learning,  \eg,  ~\cite{zhang2020dense}. Different with the existing methods to improve the SOD performance on  remote sensing images, this paper is focused on the adversarial attack and defense of the deep learning based SOD models. 
\subsection{Adversarial Attack}
There are two types of adversarial attacks: {\em white-box} attacks, where the adversary has full access to the target model, including its parameters, \ie, the model is transparent to the adversary, and {\em black-box} attacks, where the adversary has little knowledge of the target model. As the white-box attacks are usually more destructive than black-box ones in practice, the literature more focuses on the white-box attacks. Among these white-box attacks, Szegedy \etal~\cite{szegedy2013intriguing} used a box-constrained L-BFGS method to generate effective adversarial attacks for the first time. After that, the fast gradient sign method (FGSM)~\cite{goodfellow2014explaining} used the sign of the gradient to generate attacks, with $\ell_{\infty}$-norm bound. 
As a multi-step attack method, the projected gradient descent (PGD) was proposed in~\cite{madry2017towards}. Carlini and Wagner~\cite{carlini2017towards} proposed the so-called CW attack which is a margin-based attack. More recently, Croce~\etal introduced a parameter-free attack named AutoAttack~\cite{AA_atack_2020}, which is an ensemble of four diverse attacks, including two proposed variants of PGD attacks and two existing complementary attacks, \ie, FAB~\cite{croce2020minimally} and Square Attack~\cite{andriushchenko2020square}. 
Besides the perturbation ones, the attacks could also be the small geometric transformations~\cite{goodfellow2009measuring, kanbak2018geometric} or designed adversarial patches~\cite{huang2020universal, kurakin2016adversarial}.

\begin{figure*}[!]
\centering
\footnotesize
\includegraphics[width=1.0\textwidth]{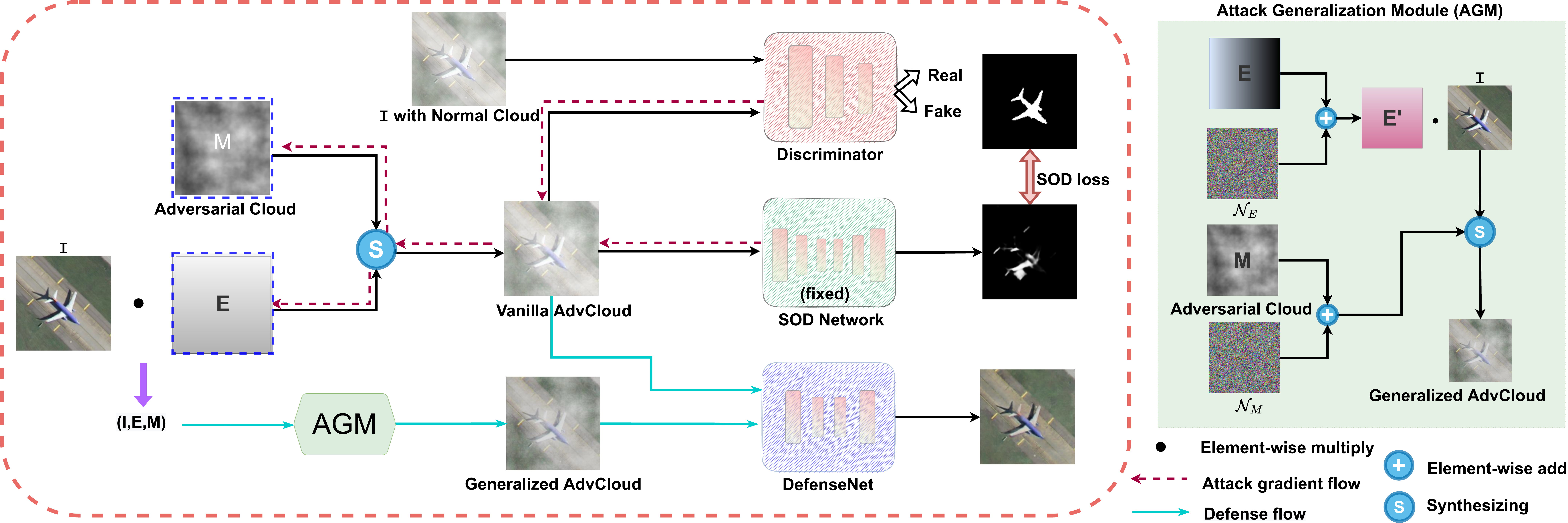}
\caption{Structure of the proposed Adversarial Cloud (AdvCloud) based attack and the proposed DefenseNet as the defense against the AdvCloud for the remote sensing Salient Object Detection (SOD). $\mathcal{N}_\text{E}, \mathcal{N}_\text{M}$ are Gaussian noises for the Attack Generalizion Module (AGM). Given a clean image $\mathbf{I}$ multipled by Exposure matrix $\mathbf{E}$ and summation of cloud mask  $\mathbf{M}$, the synthesized cloudy   image  $\hat{\mathbf{I}}$ could be obtained. The DefenseNet is a learn-able pre-processing for the SOD network.}
\label{fig:vit}	
\end{figure*}

\subsection{Adversarial Defense}
With the development of adversarial examples, studies on how to defend against those attacks and improve the robustness of the neural networks emerge. Among them, the most effective and widely used defense model is adversarial training (AT), although the most straightforward way is simply by attaching a detection network to detect and reject adversarial examples~\cite{lu2017safetynet}. AT based models, which aim to minimize the loss function to the strongest adversarial attacks within a constraint, were first proposed by~\cite{goodfellow2014explaining}. After that, a number of defending methods~\cite{madry2017towards,Mao2019,Zhong2019,shafahi2019adversarial,zhang2019theoretically,MART_2020,zhang2019defense,wang2021agkd} based on adversarial training were proposed. For example, ~\cite{Mao2019} and~\cite{Zhong2019} built a triplet loss to enforce a clean image and its corresponding adversarial example has a short distance in feature space.
TRADES~\cite{zhang2019theoretically} optimized the trade-off between robustness and accuracy.
In addition to focusing on the on-training model that utilizes adversarial examples, \cite{wang2021agkd} proposed to explore the information from the model trained on clean images by using an attention guided knowledge distillation.
Besides the adversarial training, there are also a number of other defense models have been designed. For example, Xie \etal~\cite{xie2019feature} proposed feature denoising models by adding denoise blocks into the architecture to defend the adversarial attacks, while Cohen \etal~\cite{cohen2019certified} proposed to use randomized smoothing to improve adversarial robustness. Several methods aimed to reconstruct the clean image by using a generative model~\cite{song2018pixeldefend, sun2019adversarial, yuan2020ensemble}.

\section{METHODOLOGY}


\subsection{Cloud Synthesizing for Remote Sensing}\label{subsec:cloud_syn}

Given a clean remote sensing color image $\mathbf{I}\in\mathds{R}^{H\times W \times 3}$, we aim to simulate a cloudy image via $\hat{\mathbf{I}} = \text{Cloud} (\mathbf{I}, \mathbf{E}, \mathbf{M})$, where $\mathbf{E}\in\mathds{R}^{H \times W \times 1}$ is an exposure matrix to define  exposure degree,  $\mathbf{M}\in\mathds{R}^{H\times W \times 1}$ is a cloud mask to simulate clouds,  and $\text{Cloud}(\cdot)$ represents the cloudy image synthesis function. The cloud mask $\mathbf{M}$ can be synthesized via a summation of multi-scale random noises, and is defined as 
{\small
\begin{align}\label{eq:cloud_mask}
\mathbf{M} = \sum_s \mathbf{R} \left(\mathbf{f}(2^s)\right)/2^s,
\end{align}
}\noindent
where $\mathbf{f}$ represents a randomizing function, $\mathbf{R}$ denotes a resize process and $s$ is a scale factor. $\mathbf{f}$ produces random noises with the image size $2^{s}$ followed by being resized by $\mathbf{R}$. $s$ is a natural number with range $\in$ [1, $\text{log}_{2}N$], where $N = H \times W$ is the image size. Given a clean image $\mathbf{I}$, exposure matrix $\mathbf{E}$, and cloud mask $\mathbf{M}$, we could synthesize a cloudy image $\hat{\mathbf{I}}$ via
{\small
\begin{align}\label{eq:cloud_syn}
\hat{\mathbf{I}} = \text{Cloud}(\mathbf{I}, \mathbf{E}, \mathbf{M})  
&=  \mathbf{I} \odot \mathbf{E} \odot (1 - \mathbf{M}) + \mathbf{M}, 
\end{align}
}\noindent
where $\odot$ denotes pixel-wise multiplication. 

With this cloudy image synthesis, we could study the effects of cloud from the viewpoint of adversarial attack by tuning the exposure matrix $\mathbf{E}$ and cloud mask $\mathbf{M}$ to render the synthesized cloudy images to fool the deep learning based SOD models. Later, we also employ these adversarial examples, obtained by the proposed attack method, to study the defense performance.
\subsection{Network Architecture}\label{subsec:framework}
In this section, we show the whole pipeline of adversarial cloud attack (AdvCloud), and DefenseNet as attack and defense stages to fully explore the cloud effects to a deployed deep SOD model in Fig.~\ref{fig:vit}. In the attack stage, given a clean image $\mathbf{I}$, an exposure matrix $\mathbf{E}$, a cloud mask $\mathbf{M}$, a pre-trained deep remote sensing SOD model $\phi(\cdot)$, and a well-trained discriminator $\mathcal{D}(\cdot)$, we aim to generate adversarial cloudy image examples via the proposed AdvCloud. Then, we analyze how the synthetic adversarial cloudy images hurt the SOD performance. As the other main step of the pipeline, we perform defense process, \ie, DefenseNet, as a pre-processing stage for the adversarial images to generate cloud-removed images as defense for the SOD model. The proposed DefenseNet can avoid retraining the deep SOD model and make the salient object detector process adaptive to cloudy images. For optimization, the proposed pipeline aims to maximize the detection loss of SOD model and minimize the adversarial loss of the discriminator in the attack stage to generate adversarial cloudy images which are close to normal cloudy images, while minimizing the detection loss of salient object detector by predicting a clean image in the defense stage to maintain the accuracy of the SOD model. 

\subsection{Adversarial Cloud based Attack}\label{subsec:cloud_attack}
In general, adversarial attack fails a deep model by adding an imperceptible noise-like perturbation to an image under the guidance of the deep model. In this work, we propose a novel adversarial attack method, \ie, AdvCloud, to generate adversarial cloudy remote sensing images that can fool the SOD model to verify the robustness of the SOD model. 


By intuition, we can tune $\mathbf{E}$ and $\mathbf{M}$ to generate adversarial cloudy images. Specifically, given $\mathbf{I}$,  $\mathbf{E}$, $\mathbf{M}$, and a pre-trained SOD detector $\phi(\cdot)$, we aim to tune the $\mathbf{E}$ and $\mathbf{M}$ under a norm ball constraint by  
{\small
\begin{align}\label{eq:adv_obj}
\argmax_{\mathbf{E}, \mathbf{M}}\mathcal{J}(\phi(\text{Cloud}(\mathbf{I}, \mathbf{E}, \mathbf{M})), y), \nonumber\\
\text{subject~to~~}\|\mathbf{M}-\mathbf{M}_0\|_\text{p}\leq\epsilon_\text{M},
\|\mathbf{E}-\mathbf{E}_0\|_\text{p}\leq\epsilon_\text{E},
\end{align}
}\noindent
where $\mathcal{J}(\cdot)$ is the loss function of the SOD model $\phi(\cdot)$ under the supervision of the annotation label $y$. We set $\epsilon_\text{E}$ and $\epsilon_\text{M}$ as the ball bound under $L_\text{p}$ around their initialization (\ie, $\mathbf{E}_0$ and $\mathbf{M}_0$) for the parameters $\mathbf{E}$ and $\mathbf{M}$ to avoid the clean image $\mathbf{I}$ being changed significantly. 

Similar to existing perturbation based adversarial attacks (\eg,~\cite{madry2017towards}), the object function, \ie, \reqref{eq:adv_obj}, can be optimized by gradient descent-based methods. In specific: \ding{182} We initialize $\mathbf{E}_0$ as a mask with all elements as 1 and set $\mathbf{M}_0$ via \reqref{eq:cloud_mask}. Then, we get the initial synthesized cloudy image by \reqref{eq:cloud_syn}. \ding{183} We feed the synthesized image to the SOD model $\phi(\cdot)$ and calculate the SOD loss $\ell$. \ding{184} We perform back-propagation to obtain the gradient of $\mathbf{E}$ and $\mathbf{M}$ with respective to the loss function. \ding{185} We calculate the sign of the gradient to  
update the variables $\mathbf{E}$ and $\mathbf{M}$ by multiplying the sign of their gradients with the corresponding step sizes for the next iteration, which is formulated to
{\small
\begin{align}
\label{eq:gradient_update}
\mathbf{\ell} &= \mathcal{J}(\phi(\text{Cloud}(\mathbf{I}, \mathbf{E}_{i}, \mathbf{M}_{i})), y), \nonumber\\ 
\mathbf{M}_{i+1} &= \mathbf{M}_{i} + \alpha_\text{M} \cdot \mathbf{sign}(\nabla_{\mathbf{M}_{i}} (\mathbf{\ell})), \nonumber\\ 
\mathbf{E}_{i+1} &= \mathbf{E}_{i} + \alpha_\text{E} \cdot \mathbf{sign}(\nabla_{\mathbf{E}_{i}} (\mathbf{\ell}))), 
\end{align}
}\noindent
where $\alpha_\text{M}$ and $\alpha_\text{E}$ represents the step sizes, and $i\in\{0, 1,\ldots, K-1\}$ is the iteration number. \ding{186} We generate a new adversarial cloudy image and loop from \ding{183} to \ding{185} for $K$ iterations.

\begin{figure}[h]
\centering
\footnotesize
\includegraphics[width=1.0\linewidth]{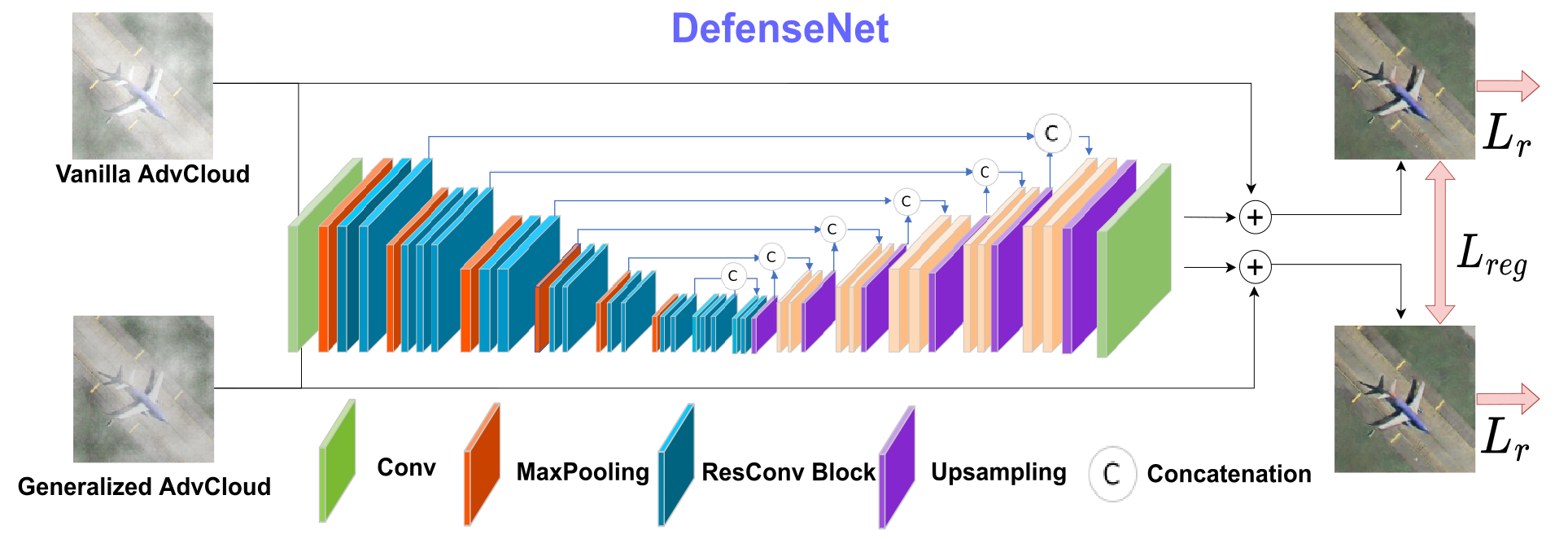}
\caption{Structure of the proposed DefenseNet.}
\label{fig:DefenseNet}	
\end{figure}

\begin{algorithm}[b]
\caption{\textbf{Defense} algorithm against the Adversarial Cloud based attack for remote sensing SOD.}
	\textbf{Input:} Clean images from the training set of EORSSD, $\epsilon_\text{M}=0.03$, $\epsilon_\text{E}=0.06$,  iteration $K=10$, $\alpha_\text{M}=0.003$, $\alpha_\text{E}=0.015$, a pre-trained remote sensing SOD detector  $\phi(\cdot)$~\cite{zhang2020dense}, and a pre-trained discriminator $\mathcal{D}(\cdot)$ obtained by pre-processing on training set. 
	\textbf{Output:} Adversarial Cloudy Images,   parameter $\theta$ for DefenseNet.
	\begin{algorithmic}[1]
		\REPEAT 
		\STATE \textit{Attack Step}:
    		\STATE
    		\begin{itemize}[noitemsep, nolistsep]
    		 \item Initial cloudy image synthesizing by \reqref{eq:cloud_syn} with $\mathbf{E_0}$ and $\mathbf{M_0}$.
    		 \end{itemize}
    		\STATE
    		\begin{itemize}[noitemsep, nolistsep]
    		\item Solve \reqref{eq:adv_obj+dis} via \reqref{eq:gradient_update_new} to obtain optimal $\mathbf{E}$ and $\mathbf{M}$ with $K$ iterations for each image to learn the corresponding adversarial cloudy image $\hat{\mathbf{I}}$.
    		\end{itemize}
		\STATE \textit{Defense Step}: 
    		\STATE
    		\begin{itemize}[noitemsep, nolistsep] 
    		\item Obtain the generalized adversarial cloudy image $\hat{\mathbf{I}}_g$ via \reqref{eq:aug_i}.
    		\end{itemize}
    		\STATE
    		\begin{itemize}[noitemsep, nolistsep]
    		\item Solve ~\reqref{eq:defense} via AdamW optimizer~\cite{loshchilov2017decoupled} to obtain optimal $\theta$ by fixed  $\mathbf{E}$ and $\mathbf{M}$ (\ie, an adversarial cloudy image $\hat{\mathbf{I}}$, the generalized adversarial cloudy image $\hat{\mathbf{I}}_g$).
            \end{itemize}
		\UNTIL{convergence or maximum epochs reached.}
	\end{algorithmic}
	\label{algo}
\end{algorithm}

\begin{figure*}[t]
\centering
\includegraphics[width=1.0\textwidth]{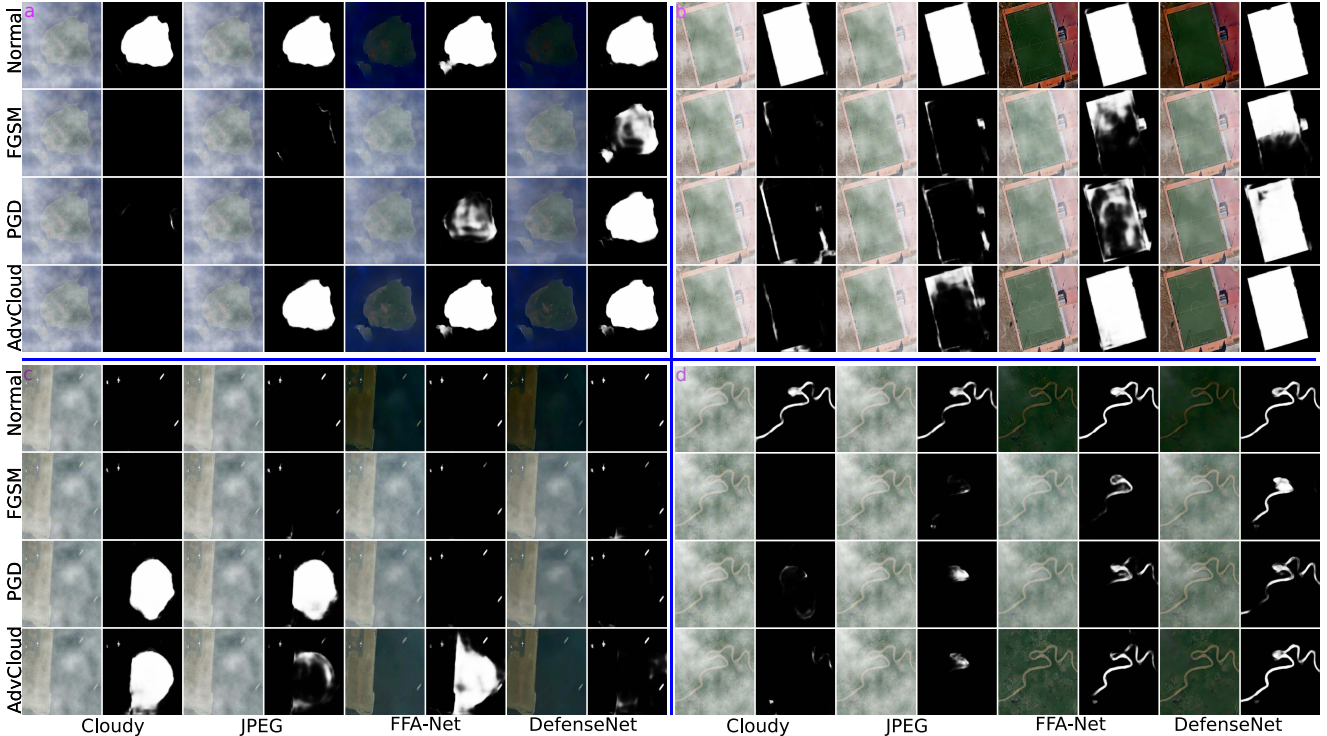}

\caption{Defense against the remote sensing salient object detection attacks. From top to bottom: normal cloudy image, attacked cloudy images by FGSM~\cite{goodfellow2014explaining}, PGD~\cite{madry2017towards}, and the proposed AdvCloud. From left to right: cloudy images, defense images by JPG Compression~\cite{das2018shield}, FFA-Net~\cite{qin2020ffa}, proposed DefenseNet, each of which is followed by its corresponding SOD result.}
\label{fig:cloud_attack}	
\end{figure*}
To make the adversarial cloudy image $\hat{\mathbf{I}}$ have close visualized  perception to the normal cloudy image, we also incorporate a discriminator $\mathcal{D}$ to align the distribution of normal cloudy images and adversarial cloudy images to avoid artifacts which might be introduced by \reqref{eq:adv_obj}. The inputs of the discriminator are an adversarial cloudy image $\hat{\mathbf{I}}$ and a normal cloudy image $\mathbf{I}_{c}$, obtained by $\mathbf{I}_c = \text{Cloud}(\mathbf{I}, \mathbf{M})  
= \mathbf{I} \odot (1 - \mathbf{M}) + \mathbf{M}$, then the adversarial training loss of the discriminator $\mathcal{D}$ is 
{\small
\begin{align}\label{eq:gan_loss}
    \mathcal{L}_{\mathcal{D}}(\hat{\mathbf{I}},\mathbf{I}_{c}) &= \mathds{E}_{\mathbf{I}_{c}\sim\mathbf{X}_{c}}[\log(\mathcal{D}(\mathbf{I}_{c}))] \nonumber \\
    &+ \mathds{E}_{\hat{\mathbf{I}}\sim\hat{\mathbf{X}}}[\log(1-\mathcal{D}(\hat{\mathbf{I}}))], 
\end{align}
}\noindent
where $\mathbf{I}_{c}$ and $\hat{\mathbf{I}}$ are instances from normal cloudy images set $\mathbf{X}_{c}$ and adversarial cloudy images set $\hat{\mathbf{X}}$, respectively.

The whole attack pipeline, incorporating AdvCloud and discriminator $\mathcal{D}$, is trained on the training set of the remote sensing SOD dataset EORSSD~\cite{zhang2020dense}. The above 
setting has an assumption for a reliable discriminator $\mathcal{D}$ ahead for the following inference stage. Specifically, we alternatively freeze adversarial parameters $\mathbf{E}$, $\mathbf{M}$ and the discriminator $\mathcal{D}$ to optimize the other one to get the reliable discriminator $\mathcal{D}$ in the training set of EORSSD$_c$ before the following inference stage.

For the inference stage of the proposed AdvCloud attack, we attack the testing set of EORSSD guided by the pre-trained discriminator $\mathcal{D}(\cdot)$ and the SOD detector $\phi(\cdot)$. Given a clean image $\mathbf{I}$ from the testing set of EORSSD, exposure matrix $\mathbf{E}$ and cloud mask $\mathbf{M}$, a well-trained discriminator $\mathcal{D}(\cdot)$, and a SOD detector $\phi(\cdot)$, we tune $\mathbf{E}$ and $\mathbf{M}$ for $K$ iterations based on back-propagation, while the optimization function \reqref{eq:adv_obj} is reformulated to 
{\small
\begin{align}\label{eq:adv_obj+dis}
\argmax_{\mathbf{E}, \mathbf{M}}(\mathcal{J}(\phi(\text{Cloud}(\mathbf{I}, \mathbf{E}, \mathbf{M})), y) - \mathcal{L}_{\mathcal{D}}(\hat{\mathbf{I}}, \mathbf{I}_c)), \nonumber\\
\text{subject~to~~}\|\mathbf{M}-\mathbf{M}_0\|_\text{p}\leq\epsilon_\text{M},
\|\mathbf{E}-\mathbf{E}_0\|_\text{p}\leq\epsilon_\text{E},
\end{align}
}\noindent
which means the adversarial cloudy image $\hat{\mathbf{I}}$ could fail the SOD detector and have the realistic cloud appearance and pattern close to normal cloudy images. Then, the updating process of variables $\mathbf{E}$ and $\mathbf{M}$, in \reqref{eq:gradient_update}, is reformulated to
{\small
\begin{align}
\label{eq:gradient_update_new}
\mathbf{\ell} &= \mathcal{J}(\phi(\text{Cloud}(\mathbf{I}, \mathbf{E}_{i}, \mathbf{M}_{i})), y), \nonumber\\ 
\mathbf{M}_{i+1} &= \mathbf{M}_{i} + \alpha_\text{M} \cdot \mathbf{sign}(\nabla_{\mathbf{M}_{i}}(\mathbf{\ell}  - \mathcal{L}_{\mathcal{D}}(\hat{\mathbf{I}}, \mathbf{I}_c))), \nonumber\\ 
\mathbf{E}_{i+1} &= \mathbf{E}_{i} +  \alpha_\text{E} \cdot \mathbf{sign}(\nabla_{\mathbf{E}_{i}}(\mathbf{\ell}  - \mathcal{L}_{\mathcal{D}}(\hat{\mathbf{I}}, \mathbf{I}_c)))).
\end{align}
}\noindent
After obtaining the updated $\mathbf{E}$ and $\mathbf{M}$ for each image from the testing set of EORSSD, we can get the corresponding adversarial cloudy images via \reqref{eq:cloud_syn}.

\subsection{Defense against Adversarial Cloud}\label{subsec:cloud_defense}
The proposed AdvCloud attack can easily hurt the SOD performance, while performing defense against adversarial attack is an effective way to alleviate such performance drop. In this section, we propose a DefenseNet as a learn-able pre-processing for adversarial cloudy images to acquire cloud-removed images for SOD models to improve the robustness. The proposed DefenseNet contains the two following branches as the inputs. 


\textbf{Vanilla AdvCloud Branch.} 
Given the updated adversarial attack variables $\mathbf{E}$ and $\mathbf{M}$, we can obtain an adversarial cloudy image $\hat{\mathbf{I}}$. Then, it is the first-branch input to the DefenseNet to perform the reconstruction for adversarial cloud removal. This is a simple white-box defense setting to make DefenseNet see the proposed AdvCloud attack so as to defend it.

\textbf{Generalized AdvCloud Branch.} 
To benefit a black-box defense making DefenseNet robust to other cloud based adversarial examples generated by different attack methods which are never seen before, we design an Attack Generalization Module (AGM) to include the generalized AdvCloud images. We use two different levels of Gaussian noise to simulate the changes produced by the gradient-based learned exposure matrix ($\mathbf{E}$) and cloud mask ($\mathbf{M}$) under a specified budget. Specifically, we add Gaussian noise $\mathcal{N}_\text{E}$ = $\omega_\text{E} \cdot \mathcal{N}(\cdot)$ and $\mathcal{N}_\text{M}$ = $\omega_\text{M} \cdot \mathcal{N}(\cdot)$ to $\mathbf{E}$ and $\mathbf{M}$ respectively to obtain  $\mathbf{E}_g$ and $\mathbf{M}_g$ so as to extend the distribution space of parameters around the gradient direction, where $\mathcal{N}(\cdot)$ is a standard Gaussian random noise generation function in the range of  [-1, 1]. Then, we could acquire a generalized  adversarial cloudy image $\hat{\mathbf{I}}_g$ with the generalized  $\mathbf{E}_g$ and $\mathbf{M}_g$ via ~\reqref{eq:cloud_syn}, \ie, 
{\small
\begin{align}\label{eq:aug_i}
\hat{\mathbf{I}}_g = \text{Cloud}(\hat{\mathbf{I}}, \mathbf{E}_g, \mathbf{M}_g), 
\end{align}
}\noindent
as the second-branch input to the DefenseNet. 

\begin{figure}[t]
\centering
\footnotesize
\includegraphics[width=1.0\linewidth]{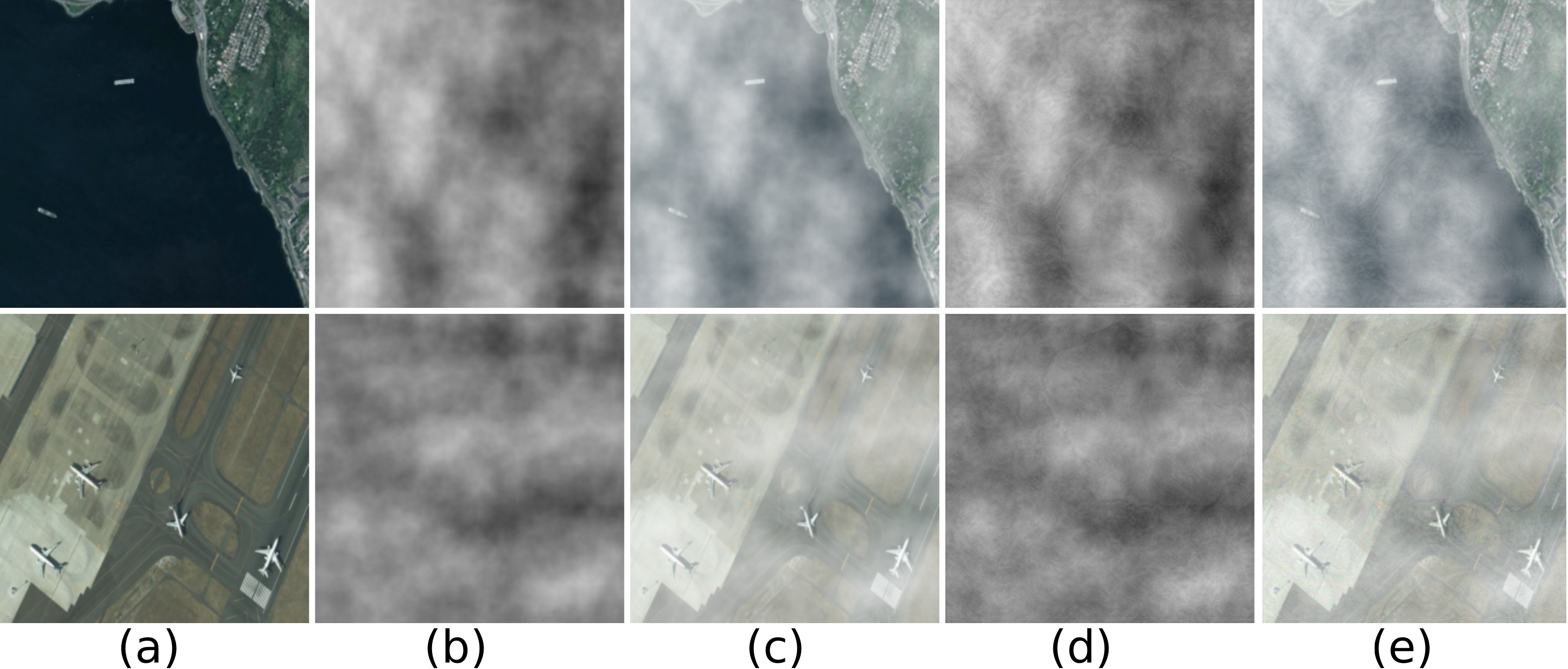}
\caption{Example images of remote sensing datasets EORSSD, EORSSD$_{c}$, EORSSD$_{adv}$. (a) clean image of EORSSD, (b) synthesized normal cloud, (c) clean image with normal cloud leading to EORSSD$_{c}$, (d) proposed adversarial cloud, and (e) clean image with proposed adversarial cloud leading to EORSSD$_{adv}$.}
\label{fig:cloud_example}
\end{figure}

\textbf{DefenseNet Loss.} 
We feed adversarial cloudy images $\hat{\mathbf{I}}$ and $\hat{\mathbf{I}}_g$ to the DefenseNet to output the cloud-removed images $\mathbf{I}^\prime = \text{DefenseNet}(\hat{\mathbf{I}};\theta)$ and $\mathbf{I}^\prime_g = \text{DefenseNet}(\hat{\mathbf{I}}_g;\theta)$, respectively. $\theta$ means the parameters of $\text{DefenseNet}$. In the defense stage, the output cloud-removed images are optimized by the image reconstruction loss function $L_r$ and regularization loss item $L_{reg}$. The object function is shown below:
{\small
\begin{align}
\label{eq:defense}
\mathcal{L} = L_r(\mathbf{I}^\prime, \mathbf{I})
+L_r(\mathbf{I}^\prime_g, \mathbf{I}) + w L_{reg}(\mathbf{I}^\prime, \mathbf{I}^\prime_g),
\end{align}
}\noindent 
where $\mathbf{I}$ is the clean image for $\hat{\mathbf{I}}$ and $\hat{\mathbf{I}}_g$, and $w$ is the balance weight which is set to 0.1. $L_r$ and $L_{reg}$ loss functions are both implemented as $L_1$ loss. 

The whole algorithm flow for the defense against the Adversarial Cloud based attack for remote sensing salient object detection is summarized in Algorithm~\ref{algo}.

\subsection{Structure of Proposed DefenseNet}
For implementation, we design the proposed DefenseNet shown in Fig.~\ref{fig:DefenseNet}. DefenseNet consists of 6 basic residual blocks, where each block includes 2 convolution layers,  one ReLu layer, and one Batch Normalization layer. The first four stages are adopted from ResNet, but the first convolution layer has 64 filters with a size of  $3 \times 3$ and stride of 1. This makes that the early feature map has the same resolution as the input image, which can lead to a bigger receptive field. There is also a bottleneck stage after the encoder part, and it consists of three convolutions layers with 512 dilated $3\times 3$  filters, and  all these convolutions layers are also followed by a batch normalization and a ReLu activation function. There is a residual block from the input to the output, making the network to focus on the residual learning.

\section{Experiments}  
\subsection{Experimental Setting}
\textbf{Benchmark Datasets:} To evaluate the salient object detection in remote sensing images, we use the public EORSSD dataset~\cite{zhang2020dense} to perform experiments. It has 2,000 remote sensing satellite images and corresponding pixel-level labeled salient object detection ground truth, which includes 1,400 images for training and 600 images for testing. The EORSSD dataset includes the objects of Aircraft,  Building,  Car,  Island,  Road,  Ship,  Water, None, and  Other in the satellite images. This dataset is quite challenging with complicated scene types, complex object attributes, comprehensive real-world satellite circumstances, and some small-size objects, therefore it is more difficult than the normal salient object detection datasets with natural images. Using each clean image in EORSSD dataset, we generate its corresponding image with the normal cloud, leading to a new synthetic dataset named EORSSD$_c$. Similarly, adding the proposed Adversarial Cloud (AdvCloud) to each clean image of EORSSD dataset, we could generate a new synthetic dataset named EORSSD$_{adv}$.  Figure~\ref{fig:cloud_example} shows some example images of the datasets EORSSD, EORSSD$_c$, and EORSSD$_{adv}$.

\begin{figure*}[t]
\centering
\footnotesize
\includegraphics[width=1.0\textwidth]{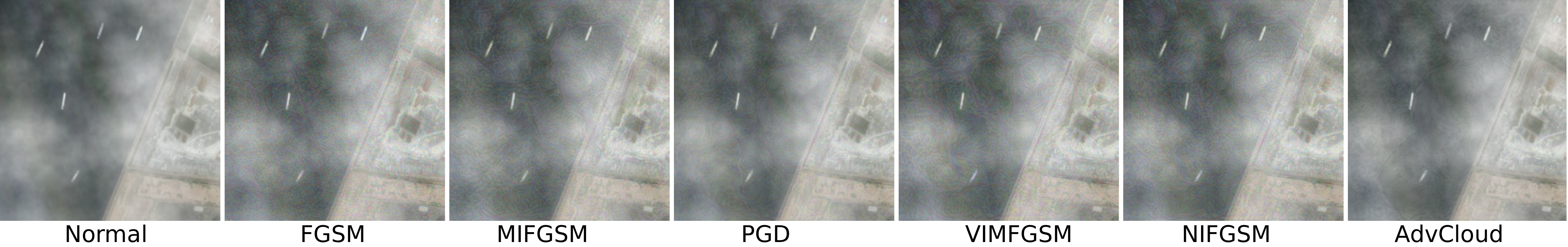}
\caption{Visualization of normal cloudy image and attacked cloudy examples by different attack methods.}
\label{fig:attack_comparsion}
\end{figure*}

\textbf{Evaluation Metrics:} We evaluate the remote sensing salient object detection performance using F-measure ($\text{F}_{\beta}$), Mean Absolute Error (MAE) score and S-measure ($\text{S}_{m}$), same as those in~\cite{zhang2020dense}. The larger F-measure, S-measure values and lower MAE score mean the better remote sensing SOD performance. Based on these metrics, we could also compare the performance of attack and defense for the remote sensing SOD task.

\textbf{Comparison Methods:}  
For the attack experiment, we compare the proposed AdvCloud  method with five additive perturbation based white-box attack methods on the EORSSD$_c$ dataset, \ie, FGSM~\cite{goodfellow2014explaining},  MIFGSM~\cite{dong2018boosting}, PGD~\cite{madry2017towards}, VMIFGSM~\cite{wang2021enhancing}, and NIFGSM~\cite{lin2019nesterov}. The maximum perturbation for these comparison methods is set to be 8 with pixel values in [0, 255]. These comparison attack methods are applied on the testing images of EORSSD$_{c}$.


\begin{table*}[ht]
\centering
\caption{Baseline remote sensing SOD performance before and after the proposed adversarial cloud (AdvCloud) attack. The budget for the perturbation cloud/noise is 8 pixels. We mark white-box attacks with * and highlight the best performance in \textcolor{red}{red}. The gray  part means the black-box attacking.  } 
\label{table:comparsion_attack_result}
\footnotesize
\resizebox{1\linewidth}{!}{
\begin{tabular}{ll|ccc|>{\columncolor[gray]{0.9}}c>{\columncolor[gray]{0.9}}c>{\columncolor[gray]{0.9}}c|>{\columncolor[gray]{0.9}}c>{\columncolor[gray]{0.9}}c>{\columncolor[gray]{0.9}}c|>{\columncolor[gray]{0.9}}c>{\columncolor[gray]{0.9}}c>{\columncolor[gray]{0.9}}c  }
\toprule
    \multirow{ 2}{*}{}&\multirow{ 2}{*}{\textbf{Attack Performance}} &\multicolumn{3}{c|}{DAFNet~\cite{zhang2020dense}} & \multicolumn{3}{c|}{ BasNet~\cite{qin2019basnet} } & \multicolumn{3}{c|}{U$^2$Net~\cite{qin2020u2}}  & \multicolumn{3}{c}{ RRNet~\cite{cong2021rrnet}} \\ 
      &  & MAE $\uparrow$ & F$_{\beta}$ $\downarrow$ & S$_{m}$  $\downarrow$ & MAE $\uparrow$  & F$_{\beta}$ $ \downarrow$  & S$_{m}$ $\downarrow$   & MAE $\uparrow$ & F$_{\beta}$  $\downarrow$ & S$_{m}$ $\downarrow$   & MAE $\uparrow$   & F$_{\beta}$  $\downarrow$ & S$_{m}$  $\downarrow$   \\       \hline

      & Clean Image & 0.0060 &0.9049& 0.9058 &0.0162& 0.8071& 0.8871  &  0.0157& 0.7890 & 0.8516 &  0.0077& 0.9086& 0.925\\
& Normal cloud & 0.0126	&0.8253	 & 0.8540	& 0.0295	& 0.7270	& 0.8352	& 0.0359	& 0.6170	& 0.7410	& 0.0100	& 0.8345	& 0.8917 \\ 
    \hline
   \multirow{ 9}{*}{ \rotatebox{90}{ATTACK}}  & FGSM  & 0.0432 * & 0.2880 * & 0.5773 * & 0.0381 & 0.5974 & 0.7488 & 0.0441 & 0.5027 & 0.6743 & 0.0202 & 0.6815 & 0.7937 \\ 

     & MIFGSM  & 0.0497 * & \textcolor{red}{0.1292} * & \textcolor{black}{0.5247} * & \textcolor{black}{0.0452} & \textcolor{black}{0.5176} & \textcolor{black}{0.7063} & \textcolor{black}{0.0461} & \textcolor{black}{0.4666} & \textcolor{black}{0.6611} & 0.0208 & \textcolor{black}{0.6344} & \textcolor{black}{0.7695} \\ 
     & PGD   & \textcolor{black}{0.0680} * & 0.1376 * & \textcolor{red}{0.5166} * & 0.0401 & 0.5860 & 0.7478 & 0.0426 & 0.5142 & 0.6869 & 0.0169 & 0.7026 & 0.8060 \\ 
     & VMIFGSM & 0.0497 * & \textcolor{black}{0.1326} * & 0.5267 * & \textcolor{red}{0.0463} & \textcolor{red}{0.4924} & \textcolor{red}{0.6952} & \textcolor{red}{0.0463} & \textcolor{red}{0.4564} & \textcolor{red}{0.6561} & \textcolor{red}{0.0245} & \textcolor{red}{0.5807} & \textcolor{red}{0.7416} \\
     & NIFGSM  & 0.0472 * & 0.1519 * & 0.5360 * & 0.0439 & 0.5176 & 0.7108 & 0.0456 & 0.4698 & 0.6623 & \textcolor{black}{0.0213} & 0.6354 & 0.7735 \\ 
     & AdvCloud w/o Noise & 0.0256 * & 0.6583 * & 0.7556 * & 0.0311 & 0.7080 & 0.8198 & 0.0373 & 0.5930 & 0.7286 & 0.0120 & 0.8018 & 0.8671 \\
     & AdvCloud w/o Exposure Matrix & 0.0484 * & 0.4265 * & 0.6435 * & 0.0317  & 0.7026 & 0.8145 & 0.0379 & 0.5953 & 0.7265 & 0.0116 & 0.8103 & 0.8765 \\ 
     &   AdvCloud  & \textcolor{red}{0.0714} * & 0.2572 * & 0.5609 * & 0.0361 & 0.6396 & 0.7771 & 0.0404 & 0.5504 & 0.7072 & 0.0143 & 0.7484 & 0.8370 \\ 

\bottomrule
\end{tabular}
}
\end{table*}

For the defense experiment, we compare our proposed DefenseNet with JPEG Compression~\cite{das2018shield}, FFA-Net~\cite{qin2020ffa}, and Defense$_{FFA}$ (using FFA-Net as the backbone). \textbf{The defense methods are all trained on EORSSD$_{adv}$ generated by attacking DAFNet which aims to remove the adversarial attack to obtain the clean image.}

For evaluating the generalization ability of the proposed attack and defense methods, we additionally employ three SOD detectors, \ie, BasNet~\cite{qin2019basnet}, U$^2$Net~\cite{qin2020u2}, and RRNet~\cite{cong2021rrnet}. All SOD models are trained on EORSSD dataset until convergence. 

Since the proposed AdvCloud are generated based on cloud, to ensure fairness in evaluating the effectiveness of different SOD (Salient Object Detection) models in attacking and defending against these adversarial examples, \textbf{the  performance  of 4 different  SOD models should treat the EORSSD$_c$ as the starting point for attacking rather than EORSSD.}

\begin{figure*}[!h]
\centering
\includegraphics[width=1\textwidth]{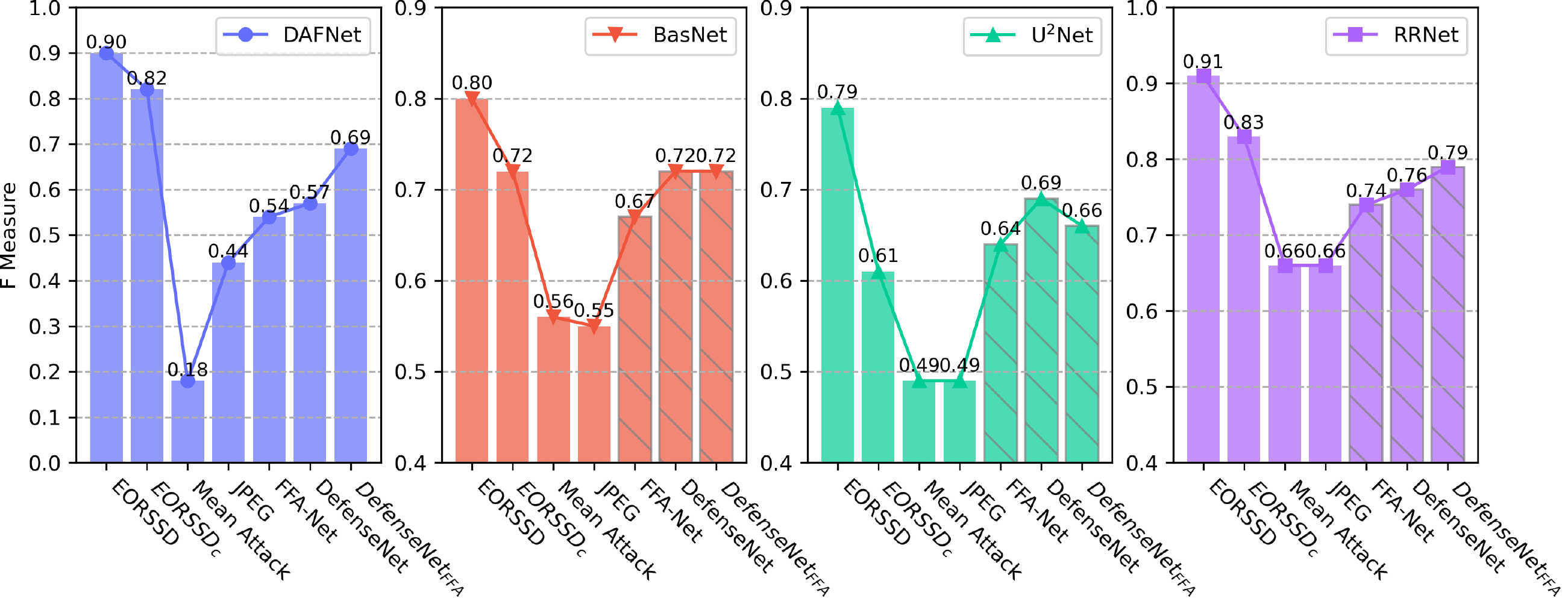}
\caption{ Visualization of defense performance across various SOD detection methods, including DAFNet, BasNet, U$^2$Net, and RRNet, in which each column represents the mean testing performance under different attack methods on  EORSSD$_c$ and defense scenarios. The EORSSD and EORSSD$_c$ represent each detector's performance under clean image and cloudy image (both without attack); the Attack column shows the mean performance under FGSM, MIFGSM, PGD, VMIFGSM, NIFGSM, and AdvCloud attacks~(Generated on DAFNet); and the subsequent columns show the mean defense results when applying JPEG, FFA-Net, DefenseNet, and Defense$_{FFA}$ methods, respectively. The gray stripes indicate the black-box defenses directly applied on attacked images without training.}
\label{fig:defense_visualization}	
\end{figure*}

\textbf{Implementation Details:} The SOD Network to be attacked is the deep learning based remote sensing salient object detection network DAFNet~\cite{zhang2020dense} 
pre-trained on the clean training images of EORSSD dataset. For the proposed AdvCloud attack, we set $\epsilon_\text{M}=0.03$, $\epsilon_\text{E}=0.06$, and the generalization random noise range of $\omega_\text{M}$,  $\omega_\text{E}$ are 0.05 and 0.1, respectively. The input image is resized to $256 \times 256$. We use the AdamW optimization algorithm~\cite{loshchilov2017decoupled} for the network training with the following hyper parameters: learning rate as 0.0001, batch size as 8, and training epoch as 80. All the experiments were run on a single NVIDIA RTX 3090 GPU card (24G). We use PyTorch to implement the proposed method.

\subsection{Experimental Results} 
\textbf{Attack Result.} 
Table~\ref{table:comparsion_attack_result} shows the quantitative SOD performance for the baseline attack. When the dataset is clean, \ie, no cloud is added, the target SOD network, DAFNet~\cite{zhang2020dense}, achieves 0.9049 overall F-measure on EORSSD dataset. After normal clouds are added to the EORSSD dataset, the F-measure decreases to 0.8253. When the proposed AdvCloud is added to the EORSSD dataset, the SOD network is misled by the adversarial examples and the F-measure is 0.2572.  This demonstrates that the proposed AdvCloud severely reduces the performance of the SOD network. Furthermore, we compare the proposed AdvCloud with other attack methods, as shown in Table~\ref{table:comparsion_attack_result}. It shows that each attack method could effectively reduce the SOD performance  Moreover, the white-box attacks on DAFNet can be effective to other SOD detectors with varying degrees of decline.

Fig.~\ref{fig:cloud_attack} shows the qualitative comparisons among different attack methods and their corresponding SOD map. Due to the attack, some objects predicted by the SOD model are ignored (a, b, d) and misidentified (c) in Fig.~\ref{fig:cloud_attack}. As we can observe, the proposed attacked image is very similar to the normal cloud in human perception compared to that from other attack methods. We can see visible defect and moire on the attacked images by other attack methods 
in Fig.~\ref{fig:attack_comparsion}. Therefore, the proposed AdvCloud is more visually close to normal cloud but with very competitive attack performance.

\begin{table*}[]
\centering
\caption{Ablation study for the defense SOD  performance of proposed DefenseNet under different attack methods. DefenseNet$^{\ddagger}$: DefenseNet w/o Generalized AdvCloud, DefenseNet$^{\dagger}$: DefenseNet w/o Vanilla AdvCloud. The white-box defense is highlighted in \textcolor{red}{red} color. }
\label{table:ablation study}
\begin{tabular}{c|ccc|ccc|ccc}
\toprule
  \multirow{2}{*}{Attack Methods} &\multicolumn{3}{c|}{DefenseNet$^{\ddagger}$} & \multicolumn{3}{c|}{DefenseNet$^{\dagger}$}  & \multicolumn{3}{c}{DefenseNet} \\
  \cline{2-10} 
   & MAE $\downarrow$ & F$_{\beta}$  $\uparrow$ & S$_{m}$  $\uparrow$  & MAE $\downarrow$ & F$_{\beta}$  $\uparrow$ & S$_{m}$  $\uparrow$   & MAE $\downarrow$   & F$_{\beta}$  $\uparrow$ & S$_{m}$  $\uparrow$   \\
\midrule
  FGSM~\cite{goodfellow2014explaining}  & 0.0373 & 0.4734 & 0.6652  & 0.0279 & 0.6161 & 0.7395  & 0.0260 & 0.6468 & 0.7548 \\
  MIFGSM~\cite{dong2018boosting} & 0.0554 & 0.3144 & 0.5966  & 0.0600 & 0.4010 & 0.6399  & 0.0569 & 0.4534 & 0.6651 \\
  PGD~\cite{madry2017towards} & 0.0400 & 0.5256 & 0.6986  & 0.0267 & 0.6770 & 0.7783  & 0.0213 & 0.7244 & 0.8039 \\
  VMIFGSM~\cite{wang2021enhancing} & 0.0659 & 0.2271 & 0.5535  & 0.0754 & 0.2844 & 0.5760  &0.0762 & 0.3268 & 0.5917 \\
  NIFGSM~\cite{lin2019nesterov} & 0.0517 & 0.3187 & 0.6004  & 0.0553 & 0.4027 & 0.6386  & 0.0516 & 0.4698 & 0.6689 \\
  Proposed AdvCloud & \textcolor{red}{0.0249} & \textcolor{red}{0.7033} & \textcolor{red}{0.8011}  & \textcolor{red}{0.0182} & \textcolor{red}{0.7477} & \textcolor{red}{0.8227}  & \textcolor{red}{0.0128} & \textcolor{red}{0.8226} & \textcolor{red}{0.8572} \\
  \midrule
  Mean &0.0459& 0.4271 &0.6526 &0.0439 &0.5215 &0.6992 &0.0408 &0.5740 &0.7236 \\
\bottomrule 
\end{tabular}
\end{table*}


\begin{table}[htbp]

\caption{Defense remote sensing SOD performance of normal cloudy images of EORSSD$_c$ with SOD detector DAFNet.}
\label{table:testing-clear} 
\resizebox{1\columnwidth}{!}{ 
\begin{tabular}{c|ccc}
\toprule
Methods & MAE $\downarrow$ & F$_{\beta}$ $\uparrow$ & S$_{m}$ $\uparrow$  \\
\hline
Clean Image & 0.0060 & 0.9049 & 0.9058 \\ 

Normal Cloud  & 0.0126 & 0.8253 & 0.8540 \\ 
\hline
JEPG Compression~\cite{das2018shield} & 0.0139 & 0.7913 & 0.8367\\
DefenseNet & 0.0171	 & 0.7747	&0.8315 \\
FFA-Net~\cite{qin2020ffa}  & 0.0144 & 0.8079 & 0.8492  \\
DefenseNet$_{FFA}$ & \textbf{0.0126} & \textbf{0.8320} & \textbf{0.8620} \\
\bottomrule
\end{tabular}}
\end{table}

\textbf{Defense Result.}
Table~\ref{tab:defense_result} shows the defense remote sensing SOD performance under different attack methods. It shows the defense methods effectively improve the SOD performance after applying defense methods to adversarial examples generated by attack strategies in  Table~\ref{table:comparsion_attack_result}.  Fig.~\ref{fig:defense_visualization} shows the comprehensive defense results on all of the attack strategies. We can clearly see that the proposed defense method, \ie,  as a  pre-processing step, achieves better F$_{\beta}$ and S$_{m}$ gains comparing with FFA-Net. The proposed DefenseNet could not only predominantly defend the proposed AdvCloud attack (\ie, white-box defense) but also effectively defend other attack methods (\ie, black-box defense). 
As shown in Table~\ref{tab:defense_result}, the F$_\beta$ performance gain by the proposed DefenseNet and DefenseNet$_{FFA}$ can be  generalization to other defense methods under each attack method. \textbf{Despite of the proposed Defense method never seen other adversarial attack images created by other attack methods during training, the proposed Defense method trained on AdvCloud can still achieve better generalization performance to defend against other attack methods}, with the help of the proposed Attack Generalization Module (AGM) shown in Table~\ref{table:ablation study}.

\begin{table*}[]
\centering
\footnotesize
\caption{ Defense performance on EORSSD$_c$ dataset. DefenseNet$_\textit{FFA}$ means the proposed DefenseNet using FFA-Net as backbone. The gray  part means the black-box defensing.}
\label{tab:defense_result}
\resizebox{1 \textwidth}{!}{ 
\begin{tabular}{ll|ccc|>{\columncolor[gray]{0.9}}c>{\columncolor[gray]{0.9}}c>{\columncolor[gray]{0.9}}c|>{\columncolor[gray]{0.9}}c>{\columncolor[gray]{0.9}}c>{\columncolor[gray]{0.9}}c|>{\columncolor[gray]{0.9}}c>{\columncolor[gray]{0.9}}c>{\columncolor[gray]{0.9}}c  }

\toprule
   \multirow{ 2}{*}{}&\multirow{ 2}{*}{\textbf{Defense Performance}} &\multicolumn{3}{c|}{DAFNet~\cite{zhang2020dense}} & \multicolumn{3}{c|}{ BasNet~\cite{qin2019basnet} } & \multicolumn{3}{c|}{U$^2$Net~\cite{qin2020u2}}  & \multicolumn{3}{c}{ RRNet~\cite{cong2021rrnet}} \\ 
      &  & MAE $\uparrow$ & F$_{\beta}$  $\downarrow$ & S$_{m}$  $\downarrow$ & MAE $\uparrow$ & F$_{\beta}$  $\downarrow$ & S$_{m}$  $\downarrow$  & MAE $\uparrow$ & F$_{\beta}$  $\downarrow$ & S$_{m}$  $\downarrow$   & MAE $\uparrow$   & F$_{\beta}$  $\downarrow$ & S$_{m}$  $\downarrow$   \\ \hline

    \multirow{ 6}{*}{ \rotatebox{90}{JPEG }}& FGSM  & 0.0332& 0.5084 &0.6756   & 0.0389& 0.5848& 0.741&0.0427& 0.5112& 0.679&0.0204& 0.6780& 0.7900 \\ 
    & MIFGSM  & 0.0421& 0.3409 &0.6117  &0.0434& 0.5147& 0.7082&0.0451& 0.4625& 0.6599& 0.0205& 0.6278& 0.7681 \\ 
     & PGD   & 0.0323&0.5205& 0.6939  &0.0396& 0.5851& 0.7479&0.0418& 0.5136& 0.6874&0.017& 0.6874& 0.8049\\ 
    & VMIFGSM & 0.0485 &0.2710& 0.5745  & 0.0464& 0.4868& 0.6926&0.0459& 0.4518& 0.6531&0.0242& 0.5818& 0.7413 \\ 
     & NIFGSM  & 0.0422& 0.3575& 0.6139  &  0.0433& 0.5211& 0.7088& 0.0447& 0.4671& 0.6628& 0.0215& 0.6262& 0.7685\\ 
     &  AdvCloud & 0.0242& 0.6228 &0.7486 & 0.0363& 0.6334& 0.7756&0.0401& 0.5524& 0.706& 0.0144& 0.7353& 0.8312 \\
        \hline

    \multirow{ 6}{*}{ \rotatebox{90}{DefenseNet }}& FGSM &0.0260 &0.6468 &0.7548 & 0.025& 0.7182& 0.8247& 0.0237& 0.6776& 0.7865& 0.0159& 0.7668& 0.8414 \\ 
    & MIFGSM  & 0.0569 &0.4534& 0.6651& 0.0265& 0.7017& 0.8154& 0.0237& 0.6776& 0.7865& 0.016& 0.7476& 0.8352 \\ 
     & PGD   & 0.0213 &0.7244& 0.8039 &0.0265& 0.7017& 0.8154 &0.0209& 0.7210& 0.8106 & 0.0126& 0.7963& 0.8656\\ 
    & VMIFGSM &  0.0762 &0.3268 &0.5917& 0.0302& 0.6689& 0.7954 &0.026& 0.6523& 0.7704& 0.0194& 0.6912& 0.8040 \\ 
     & NIFGSM  & 0.0516& 0.4698 &0.6689 & 0.0165& 0.7508& 0.8345 &0.0241& 0.6762& 0.7844 &0.0165& 0.7508& 0.8345\\ 
     &  AdvCloud & \textcolor{red}{0.0128}& \textcolor{red}{0.8226} & \textcolor{red}{0.8572}& 0.0193& 0.7496& 0.8549& 0.0173& 0.7644& 0.8368&   0.0111& 0.8365& 0.8952 \\
        \hline

    \multirow{ 6}{*}{ \rotatebox{90}{FFANet }}& FGSM  & 0.0292  & 0.5993  & 0.7309  & 0.0363 & 0.6260 & 0.7725 & 0.0369 & 0.5846 & 0.7264 & 0.0190 & 0.7015 & 0.8092 \\ 
    & MIFGSM  & 0.0535  & 0.4077  & 0.6427  & 0.0331 & 0.6607 & 0.7907 & 0.0322 & 0.6168 & 0.7485 & 0.0174 & 0.7185 & 0.8170 \\ 
     & PGD   & 0.0244  & 0.6861  & 0.7799  & 0.0332 & 0.6557 & 0.7873 & 0.0306 & 0.6439 & 0.7653 & 0.0139 & 0.7722 & 0.8529 \\ 
    & VMIFGSM & 0.0692 & 0.3017 & 0.5838  & 0.0354 & 0.6280 & 0.7711 & 0.0334 & 0.5972 & 0.7367 & 0.0205 & 0.6722 & 0.7909 \\ 
     & NIFGSM  & 0.0484  & 0.4318  & 0.6518 & 0.0332 & 0.6557 & 0.7873 & 0.0330 & 0.6112 & 0.7441 & 0.0180 & 0.7210 & 0.8182 \\ 
     &  AdvCloud & \textcolor{black}{0.0145}  & \textcolor{black}{0.7965}  & \textcolor{black}{0.8443}  & \textcolor{black}{0.0180} & \textcolor{black}{0.7826} & \textcolor{black}{0.8710} & \textcolor{red}{0.0165} & \textcolor{black}{0.7768} & \textcolor{black}{0.8462} & \textcolor{black}{0.0102} & \textcolor{black}{0.8423} & \textcolor{black}{0.8971} \\
        \hline

    \multirow{ 6}{*}{ \rotatebox{90}{Defense$_\textit{FFA}$ }} & FGSM  & 0.0224  & 0.6995  & 0.7821  & 0.0316 & 0.6891 & 0.8095 & 0.0354 & 0.6072 & 0.7363 & 0.0136 & 0.7901 & 0.8561 \\ 
    & MIFGSM  & 0.0255  & 0.6488  & 0.7618  & 0.0279 & 0.7145 & 0.8244 & 0.0301 & 0.6479 & 0.7637 & 0.0138 & 0.7788 & 0.8551 \\ 
    & PGD   & 0.0149  & 0.7778  & 0.8313  & 0.0260 & 0.7294 & 0.8357 & 0.0288 & 0.6689 & 0.7781 & 0.0122 & 0.7971 & 0.8692 \\ 
    & VMIFGSM & 0.0393  & 0.5338  & 0.6962  & 0.0291 & 0.6894 & 0.8113 & 0.0313 & 0.6278 & 0.7528 & 0.0159 & 0.7419 & 0.8306 \\ 
     & NIFGSM  & 0.0259  & 0.6512  & 0.7586  & 0.0276 & 0.7097 & 0.8227 & 0.0313 & 0.6378 & 0.7580 & 0.0139 & 0.7860 & 0.8573 \\
    &   AdvCloud & 0.0130  & 0.8178  & 0.8592  & \textcolor{red}{0.0171} & \textcolor{red}{0.7924} & \textcolor{red}{0.8761} & \textcolor{black}{0.0169} & \textcolor{red}{0.7834} & \textcolor{red}{0.8486} & \textcolor{red}{0.0097} & \textcolor{red}{0.8586} & \textcolor{red}{0.9031} \\ 
 \bottomrule    
\end{tabular}}
\end{table*}

\textbf{Ablation Study for Proposed DefenseNet.}
The proposed DefenseNet has two input branches, \ie, regular attack image branch and generalized attack image branch. Table~\ref{table:ablation study} shows both the regular attack branch and the generalized attack  branch contribute to the final defense SOD performance, where the best defense performance is obtained when combining the two branches. If the branch of generalized attack is removed, it will lead to more significant defense performance drop. The DefenseNet contain AGM module cane provide a promising and effective solution for generative defense on different adversarial attacks.

\begin{table}[!h]

\caption{Image quality comparison of different cloudy attack methods with DAFNet as the SOD detector. EORSSD$_C$: normal cloudy images, EORSSD: original clean images.}

\resizebox{1\columnwidth}{!}{ 
\begin{tabular}{l|ccc|ccc}
\toprule
\centering
 \multirow{ 2}{*}{Methods} &\multicolumn{3}{c|}{Compare with EORSSD$_C$} & \multicolumn{3}{c}{Compare with EORSSD}  \\
 
 & SSIM$\uparrow$ & PSNR$\uparrow$ & L2$\downarrow$ & SSIM$\uparrow$ & PSNR$\uparrow$ & L2$\downarrow$  \\  
\midrule

        Normal Cloud & 1 & - & 0.00 & 0.64 & 10.01 & 331.85  \\ \hline
        FGSM & 0.63 & 30.25 & 181.57  & 0.44 & 9.96 & \textbf{330.46}  \\ 
        MIFGSM & 0.70 & 31.45 & 137.95 & 0.47 & 9.99 & 330.87  \\ 
        PGD & 0.79 & 33.54 & 85.49  & 0.53 & 9.99 & 331.15  \\ 
        VMIFGSM & 0.70 & 31.37 & 121.55 & 0.47 & 9.99 & 330.79 \\ 
        NIFGSM & 0.69 & 31.24  & 137.26 & 0.47 & 9.99 & 330.78 \\ 
        ADvCloud & \textbf{0.88} & \textbf{36.24} & \textbf{46.91}  & \textbf{0.58} & \textbf{10.00} & 331.32  \\ 
\bottomrule
    \end{tabular}}
\label{table:difference}
\end{table}

\textbf{Discussion about Defense on Normal Cloudy Images.}

The DefenseNet$_{FFA}$'s performance in defense remote sensing SOD was assessed using normal cloudy images of EORSSD$_c$. The results in Table~\ref{table:testing-clear} indicate that the proposed defense mechanism is capable of effectively defending against anonymous types of attacks, while maintaining strong performance on normal images. This suggests that our defense method is reliable and effective in both attack and non-attack scenarios.

\textbf{Discussion about Visual quality.}
The image quality comparison results are shown in Table~\ref{table:difference}. It turns out that the proposed AdvCloud has better image quality after attack. We use 8-pixel as the budget for the perturbation attack noise $M$, same as all of the comparison methods. Combing with the observation in Fig.~\ref{fig:cloud_attack}, although our proposed attack method can not achieve the best attack performance, our AdvCloud attack is more imperceptible comparing with other attack methods.

\section{Conclusion}
In this paper, we proposed a new Adversarial Cloud to attack the deep learning based remote sensing salient object detection model, meanwhile a new DefenseNet as pre-processing defense is proposed to purify the input image without tuning the deployed remote sensing deep SOD model. To study this research problem, we synthesized new benchmarks EORSSD$_{c}$ with normal cloud and EORSSD$_{adv}$ with the proposed adversarial cloud from the existing remote sensing SOD dataset EORSSD. The extensive experiment on 4 SOD networks shows that the proposed DefenseNet could well pre-process the attacked cloudy images as defense against different adversarial attack methods without changing the deployed remote sensing deep SOD model, while the SOD performance on the remote sensing normal cloudy images without attack is still promising.   
\clearpage

{\small
\bibliographystyle{ieeetr}
\bibliography{main}
}

\end{document}